\PassOptionsToPackage{dvipsnames}{xcolor}
\documentclass{template/bytedance}

\usepackage[utf8]{inputenc} % allow utf-8 input
\usepackage{amsfonts}       % blackboard math symbols
\usepackage{nicefrac}       % compact symbols for 1/2, etc.

\usepackage{amsmath} 
\usepackage{enumerate}
\usepackage{enumitem}
\usepackage{wrapfig}
\usepackage{graphicx}
\usepackage{rotating}
\usepackage{colortbl}
\usepackage{float}
\usepackage{placeins}
\usepackage{pgfplots}
\pgfplotsset{compat=1.16}
\usetikzlibrary{arrows.meta,positioning,shapes.geometric,calc,fadings,decorations.pathmorphing}

\newcommand{\TheName}{DirectorBench}

\fancypagestyle{firststyle}{%
  \fancyhf{}%
  \fancyhead[L]{\vskip 4mm\includegraphics[height=8mm]{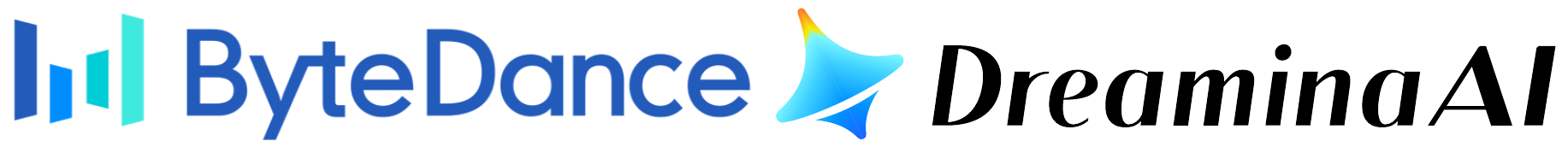}}%
}

\definecolor{autoskillblue}{RGB}{220,232,250}   % light ByteDance blue, table/figure highlight
\definecolor{autoskilldeep}{RGB}{44,90,181}     % ByteDance corporate blue (#2C5AB5)
\definecolor{autoskilldark}{RGB}{24,55,130}     % deeper navy, borders / emphasis

\definecolor{codexamber}{RGB}{224,139,60}
\definecolor{codexamberlight}{RGB}{255,221,184}
\definecolor{codexamberdark}{RGB}{162,93,28}
\definecolor{bdteal}{RGB}{69,191,199}
\definecolor{bdteallight}{RGB}{198,233,235}
\definecolor{bdtealdark}{RGB}{36,128,134}
\definecolor{deltagreen}{RGB}{22,128,57}

\usepackage{pifont}
\usepackage{wasysym}

\definecolor{markgreen}{RGB}{34,139,52}
\definecolor{markred}{RGB}{200,40,40}
\definecolor{markamber}{RGB}{210,150,0}

\lstdefinestyle{jsonstyle}{
    basicstyle=\ttfamily\footnotesize,
    breaklines=true,
    breakatwhitespace=false,
    columns=fullflexible,
    keepspaces=true,
    frame=single,
    framesep=4pt,
    xleftmargin=0pt,
    xrightmargin=0pt,
    linewidth=\linewidth,
    postbreak=\mbox{\textcolor{gray}{$\hookrightarrow$}\space}
}
\title{\TheName: Diagnosing Long-Form Video Generation with Personalized Multi-Agent Evaluation}

\author[1,2]{Jiamin Chen}
\author[1]{Qianben Chen}
\author[1]{Jiawen Zhang}
\author[1]{Yidi Wu}
\author[2]{Yuchen Li}
\author[2]{Xiaokun Zhang}
\author[1,\dagger]{Wangchunshu Zhou}
\author[2]{Chen Ma}

\affiliation[1]{ByteDance Inc.}

\affiliation[2]{City University of Hong Kong}

\contribution[\dagger]{Corresponding author}

\abstract{
Long-form video generation is rapidly moving from short, single-scene synthesis toward minute-long, multi-shot creation with narrative structure, cinematic control, audio, and cross-modal synchronization. However, evaluating such videos remains challenging, since existing benchmarks largely focus on local visual quality, short-horizon temporal consistency, or generic prompt alignment, and provide limited diagnosis of workflow failures and user-dependent preferences.
We introduce \TheName{}, a personalized multi-agent diagnostic benchmark for long-form video generation. \TheName{} evaluates generated videos with respect to 80 structured metadata entries, 7 user profiles, and 40 checkpoint criteria across 5 dimensions: script, visual, audio, cross-modal, and stability. Instead of reducing quality to a single aggregate score, \TheName{} localizes checkpoint-level bottlenecks and supports profile-aware evaluation.
We evaluate 4 long-form video generation workflows, 6 base LLMs, and 7 user profiles. Across workflows, \TheName{} reveals a between-unit bottleneck: transition quality averages only 0.256 and reaches 0.356 for the best workflow, while prompt-level user demand fulfillment averages 0.71. We further conduct human evaluation with 14 annotators to validate the alignment between \TheName{} and human judgment. The results show that \TheName{} captures human-perceptible quality differences and reveals workflow- and profile-dependent failure modes that are hidden by aggregate scoring. These findings highlight the importance of diagnostic and profile-aware benchmarking for long-form video generation.
}

\date{\today}
\github{\url{https://github.com/jiaminchen-1031/DirectorBench}}
\hf{\url{https://huggingface.co/datasets/Jiamin1031/DirectorBench}}
\correspondence{Wangchunshu Zhou, \email{chunshu@bytedance.com}}

\begin{document}

\maketitle

\begin{figure}[t]
    \centering
    \vspace{-10pt}
    \includegraphics[width=0.9\linewidth]{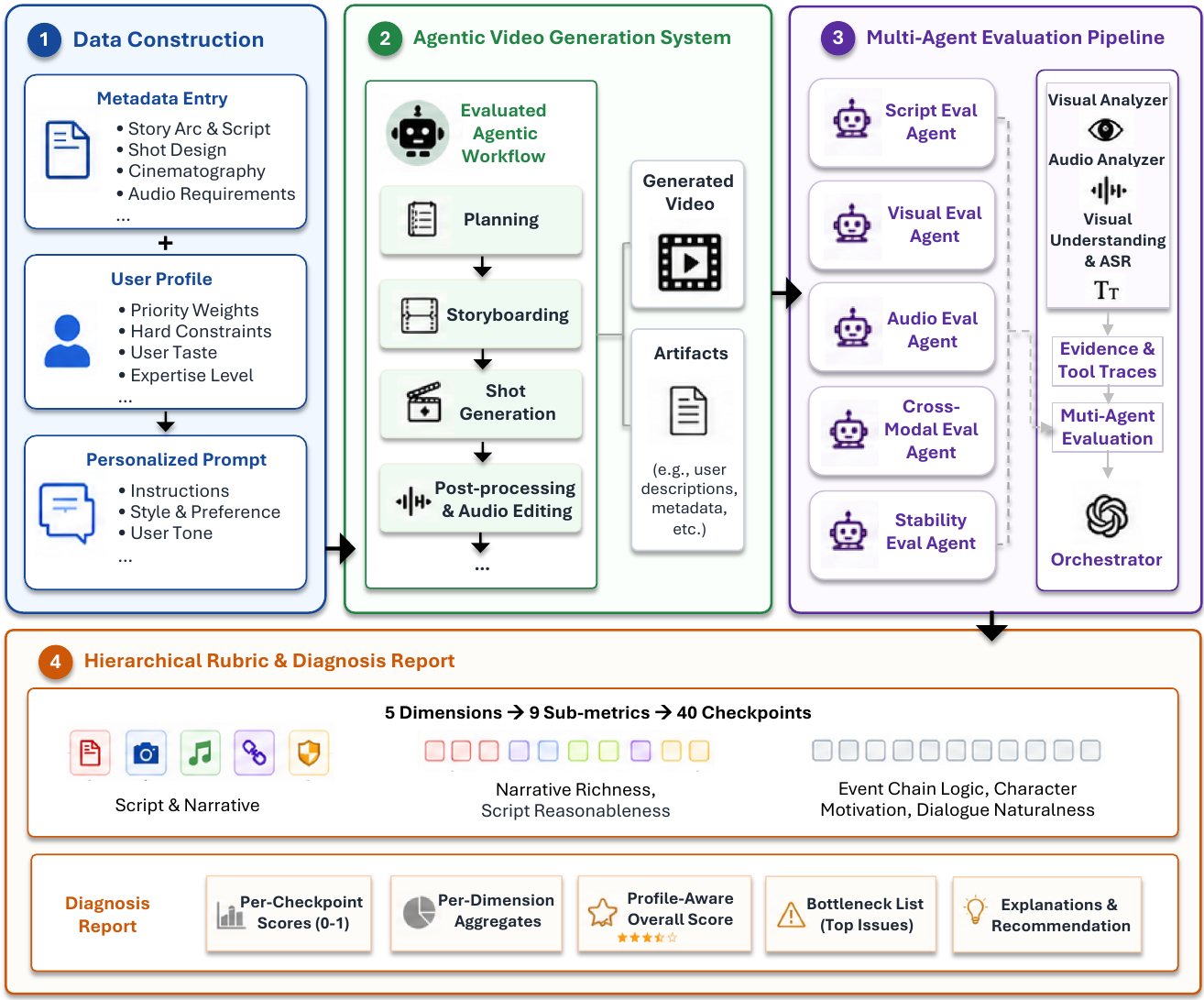}
     % \vspace{-10pt}
    \caption{Overview of DirectorBench. Metadata and user profiles are transformed into personalized prompts to query an agentic video generation system. The generated outputs are evaluated through a multi-agent pipeline and a hierarchical checkpoint rubric to produce structured diagnosis reports.}
     \vspace{-10pt}
    \label{fig:placeholder}
\end{figure}
\section{Introduction}

Recent advances in text-to-video generation~\citep{villegas2023makeavideo} have significantly improved
the visual fidelity, motion realism, and prompt controllability of short
video clips. Modern systems can generate visually appealing scenes from
natural-language prompts, and a growing ecosystem of agentic workflows
has begun to decompose video creation into stages~\citep{wu2025automated, hkuds2025vimax, meng2025holocine}. These
developments mark a shift from isolated clip generation~\citep{brooks2024video, peng2025opensora} toward
long-form, multi-shot video creation~\citep{meng2025holocine, wu2025automated}, where the goal is no longer merely
to synthesize a plausible short scene, but to produce a coherent video
with narrative structure, cinematic control, audio-visual alignment, and
long-term consistency.
% A short clip typically contains a single action, a single
% scene, or a pseudo-continuous camera movement. In contrast, long-form
% video requires multiple shots, controllable transitions, consistent
% characters and scenes, director-level camera
% language, etc. These requirements introduce failure modes that are
% largely invisible in short-clip evaluation.
As a result, long-form video quality depends not only on local frame
fidelity, but also on the structure that connects shots, modalities, and
user intent over time.

Existing video-generation benchmarks~\citep{huang2024vbench, zheng2025vbench2, han2025videobench} have made important
progress in measuring visual quality, temporal smoothness, subject
consistency, and prompt alignment. However, they remain insufficient for
evaluating long-form, agentically generated videos. Most existing
metrics are designed for short clips or frame-level properties, and
therefore under-represent between-shot failures such as transition
quality, long-term continuity, and story-level coherence~\citep{huang2024vbench, zheng2025vbench2}. In addition,
current evaluations focus primarily on visual consistency,
while audio quality, narration, and cross-modal coordination are less
systematically diagnosed~\citep{han2025videobench}. Furthermore, most benchmarks collapse quality
into a single aggregate score, ignoring that video creation is inherently
user-dependent~\citep{zheng2025vbench2, ma2026personalizedrewardbench, liu2024dreambenchpp}. At the same time, modern video-generation systems are
increasingly agentic, involving base LLMs, planning modules, storyboard
generation, tool calls, and iterative refinement~\citep{wu2025automated, hkuds2025vimax, meng2025holocine, jang2024mavis}, where failures are
distributed across components rather than attributable to a single model.
As a result, a useful benchmark should go beyond final-score comparison
and provide diagnostic signals that localize bottlenecks and explain how
different parts of the workflow contribute to overall quality.

To address these challenges, we propose \TheName{}, a personalized
multi-agent diagnostic benchmark for long-form video generation.
% \TheName{} evaluates generated videos with respect to three inputs: a
% structured metadata specification that defines the objective video
% intent, a user profile that encodes subjective priorities and hard
% constraints~\citep{ma2026personalizedrewardbench, liu2024dreambenchpp, peng2025personalizedvideo, xing2026lumosx}, and
% personalized prompts generated from each metadata--profile pair. 
\TheName{} evaluates agentic long-form video generation systems under
structured metadata specifications and personalized user profiles,
analyzing the generated outputs to produce diagnostic evaluation
signals.
The key design
principle of \TheName{} is to separate objective intent from personalized
preference: metadata specifies what the video should contain (e.g.,
story arcs, shot design, camera control, consistency constraints, and
audio requirements), while user profiles determine how different
dimensions are prioritized~\citep{ma2026personalizedrewardbench, liu2024dreambenchpp, peng2025personalizedvideo, xing2026lumosx}. This enables \TheName{} to jointly assess
whether a video satisfies concrete requirements and how its perceived
quality varies under different user preferences~\citep{ma2026personalizedrewardbench, liu2024dreambenchpp, han2025videobench, zheng2025vbench2}.

We instantiate \TheName{} with 80 structured metadata entries, 7 user
profiles, and 40 checkpoint criteria across five dimensions: script,
visual, audio, cross-modal, and stability. A dynamic checkpoint rubric
activates only criteria applicable to each video, and a multi-agent
evaluation pipeline combines specialist agents, tool-based evidence,
confidence-aware aggregation, and profile-weighted scoring. Instead of
collapsing quality into a single aggregate score, \TheName{} produces
checkpoint-level diagnosis reports that localize bottlenecks across
dimensions.
Using \TheName{}, we evaluate long-form video generation across
% agentic workflows, base LLMs, and user profiles 
various settings
to analyze how workflow design,
model choice, and personalization affect generation quality. Across
workflows, \TheName{} reveals a between-unit bottleneck: transition
quality averages only 0.256 and reaches 0.356 for the best workflow,
while prompt-level user demand fulfillment averages 0.71. We further
conduct human evaluation with multiple annotators and show that
\TheName{} aligns with human judgment at the dimension level. These
results demonstrate that diagnostic and profile-aware evaluation reveals
workflow failures and user-dependent quality variation that are hidden by
aggregate metrics.

Our contributions are threefold:
\begin{itemize}[leftmargin=*]\setlength\itemsep{-0.1em}
    \item We introduce \TheName{}, a personalized multi-agent diagnostic
    benchmark for long-form video generation, instantiated with 80
    structured metadata entries, 7 user profiles, and 40 checkpoint
    criteria across script, visual, audio, cross-modal, and stability
    dimensions.

    \item We propose a multi-agent evaluation pipeline with dynamic
    checkpoint activation, tool-based evidence, confidence-aware
    aggregation, and profile-weighted scoring, producing checkpoint-level
    diagnosis reports rather than only aggregate scores.

    \item We evaluate long-form video generation across agentic workflows,
    base LLMs, and user profiles, showing that \TheName{} reveals
    between-unit bottlenecks, workflow-specific failure signatures, and
    user-dependent quality variation, with validation from human
    evaluation.
\end{itemize}

\section{\TheName: Evaluating Long-Form Video Generation through Personalized Multi-Agent Diagnosis}\label{sec:method}
In this section, we present \TheName{}, a diagnostic benchmark for evaluating long-form, multi-shot video generation through personalized multi-agent assessment.
Rather than reducing video quality only to a single aggregate score, \TheName{} also produces a structured \emph{diagnosis report} that
(i)~reflects the evaluation perspective of a specific user profile,
(ii)~operates at checkpoint-level granularity across different modalities, and
(iii)~localizes concrete bottlenecks to guide targeted method improvement.
Figure~\ref{fig:placeholder} illustrates the overall framework.
 
%------------------------------------------------------------------
\subsection{Problem Formulation}
\label{sec:formulation}
 
% We formulate the evaluation of a long-form generated video as a function of three inputs:
 
% \begin{equation}
%     f(\mathbf{m}, \mathbf{u}, \mathbf{v}) \;\rightarrow\; \mathcal{R}
% \end{equation}
 
% \noindent where $\mathbf{m}$ is a \emph{metadata} specification encoding the objective requirements of the target video (story arc, shot design, camera directives, audio requirements, etc.), $\mathbf{u}$ is a \emph{user profile} encoding personal preferences through priority weights and hard constraints, $\mathbf{v}$ is the generated video under evaluation, and $\mathcal{R}$ is a \emph{diagnosis report} containing per-checkpoint scores, per-dimension aggregates, a profile-weighted overall score, and an explicit bottleneck list.
We formulate long-form video evaluation as the assessment of an
agentic generation system conditioned on structured metadata and user
preferences:
\begin{equation}
    f(\mathbf{m}, \mathbf{u}, \mathcal{G}) \;\rightarrow\; \mathcal{R},
\end{equation}
where $\mathbf{m}$ is a \emph{metadata} specification encoding the
objective requirements of the target video (e.g., story arc, shot
design, camera directives, audio requirements, and consistency
constraints), $\mathbf{u}$ is a \emph{user profile} encoding subjective
preferences through priority weights and hard constraints, and
$\mathcal{G}$ denotes the agentic video-generation system under
evaluation. Executing $\mathcal{G}$ conditioned on $(\mathbf{m},
\mathbf{u})$ produces one or more generated videos. The outputs are analyzed by \TheName{} to
produce a structured \emph{diagnosis report} $\mathcal{R}$.
% containing per-checkpoint scores, per-dimension aggregates, a profile-weighted overall score, and an explicit bottleneck list.
 
Personalized preferences are \emph{orthogonal} to general quality, especially in content creation: the same output can be preferred under one user profile and rejected under another, even when both exhibit
comparable general quality. In the video domain, this divergence is amplified by the multi-modal nature of the medium.
% The same generated video may receive a high score from a \emph{Story-First} user who prioritizes narrative coherence, yet a low score from a \emph{Visual-Heavy} user who demands precise cinematic control. 
This is precisely why our formulation includes the user profile $\mathbf{u}$ as an explicit input: without it, such preference-driven divergences are collapsed into a single ranking that misrepresents quality for all but an implicit ``average'' user.
Crucially, the metadata $\mathbf{m}$ serves as the \emph{ground-truth intent}, providing objective reference points for factual checkpoints. This separates \TheName{} from purely preference-based evaluation: checkpoints are anchored to verifiable specifications, while the \emph{weighting} of those checkpoints is personalized.
 
%------------------------------------------------------------------

\subsection{Benchmark Construction}
\label{sec:data}
 
The benchmark comprises three carefully designed components: metadata specifications that define \emph{what} to evaluate, user profiles that define \emph{from whose perspective} to evaluate, and personalized prompts that serve as the input to generation systems.

\begin{wrapfigure}{r}{0.5\textwidth}
    \centering
    \includegraphics[width=\linewidth]{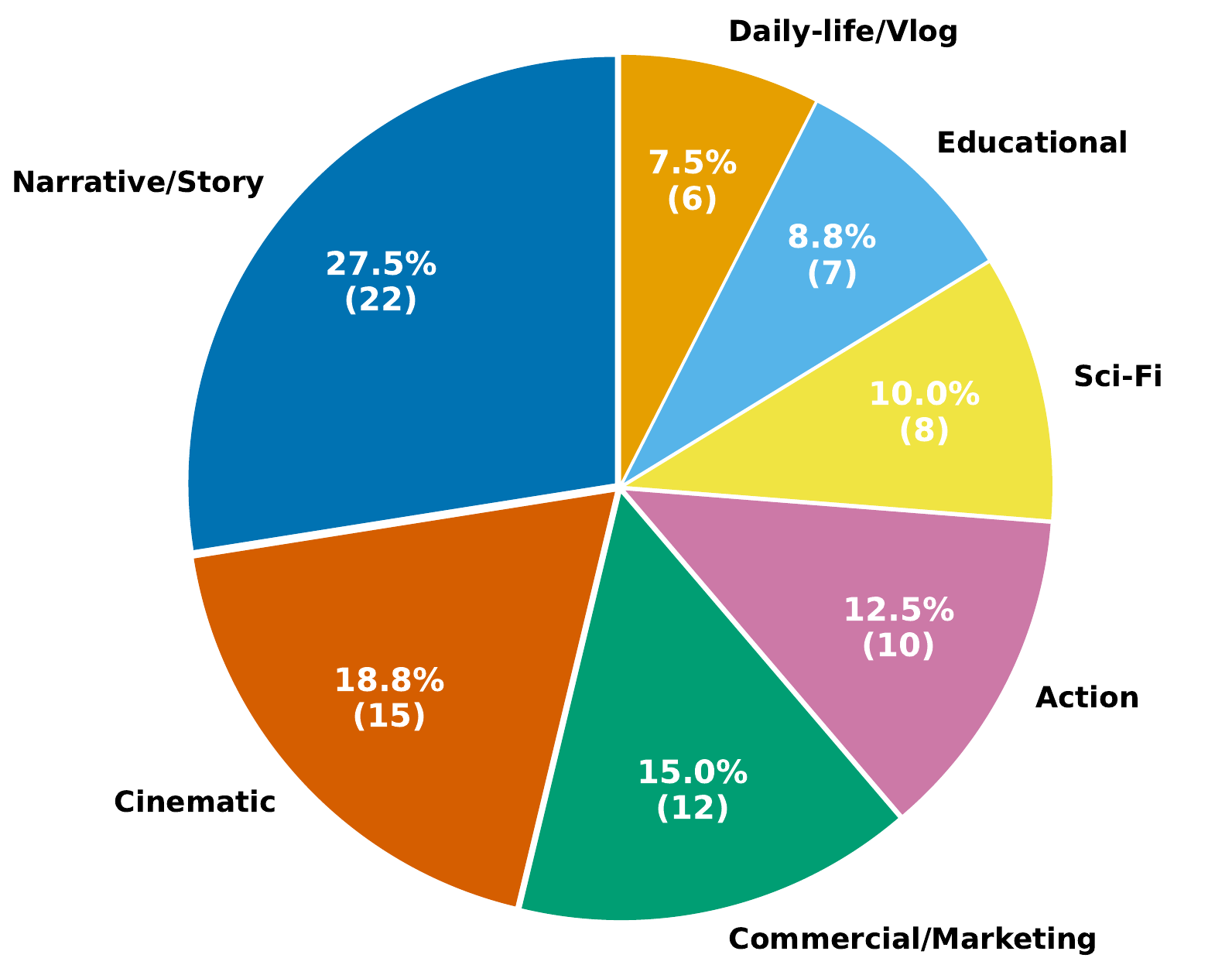}
    \caption{Distribution of the metadata entries across the various video categories.}
    \vspace{-5pt}
    \label{fig:metadata_distribution}
\end{wrapfigure}
%--- Metadata ---
\noindent\textbf{Metadata Design.}
Each metadata entry $\mathbf{m}$ encodes the complete task specification of a target video.
We design 80 metadata entries spanning 7 categories: narrative/story, commercial/marketing, educational, sci-fi, cinematic, action, and daily-life/vlog. Each entry is derived from: (i)~curated and restructured from online data, or (ii)~hand-crafted to target specific evaluation challenges. The category distribution and full breakdown are provided in Figure~\ref{fig:metadata_distribution}.
The distribution reflects both relative complexity and real-world prevalence, with narrative videos receiving the largest share due to their demanding multi-act structure.
 
Every entry is a structured JSON document specifying a unique identifier, target duration, video type, and a high-level creative instruction, alongside a detailed modality specification block covering three dimensions. The \textbf{text} dimension decomposes the narrative into a three-act story arc (setup, conflict, resolution) with per-shot script entries including dialogue, narration, and timing. The \textbf{visual} dimension defines per-shot descriptions with camera movement types, lighting conditions, and consistency requirements across character identity, clothing, scene setting, and overall style. The \textbf{audio} dimension specifies dialogue and lip-sync requirements, background music style, sound effects, tone control directives, and optional multi-language configurations.
This tri-modal structure ensures that every evaluation checkpoint has a concrete, verifiable reference, with a complete metadata example provided in Appendix~\ref{app:metadata_example}.

%  \begin{wrapfigure}{r}{0.5\textwidth}
%     \centering
%     \includegraphics[width=\linewidth]{metadata_pie_labeled.pdf}
%     \caption{Distribution of the metadata entries across the various video categories.}
%     \vspace{-5pt}
%     \label{fig:metadata_distribution}
% \end{wrapfigure}
%--- User Profiles ---
\noindent\textbf{User Profile Design.}
We define 7 user profiles $\{\mathbf{u}_1, \ldots, \mathbf{u}_7\}$ representing distinct preferences observed in real-world video creation. Each profile $\mathbf{u}$ consists of three components: (i)~\textbf{priority weights} $\mathbf{w} \in \mathbb{R}^4$ ($\sum_d w_d = 1$) 
% distributing emphasis across four dimensions including narrative quality, visual cinematography, audio emotion, and cross-modal synchronization; 
distributing emphasis across four user-facing dimensions including narrative quality, visual cinematography, audio emotion, and cross-modal synchronization; Note that the
\emph{generation stability} dimension is treated separately from
personalized preference weighting because it reflects system-level
generation reliability rather than subjective user preference. 
(ii)~\textbf{hard constraints} specifying non-negotiable requirements (e.g., perfect lip-sync, complete three-act arc) whose violation triggers a significant penalty regardless of other dimensions; and (iii)~\textbf{user taste} encoding fine-grained aesthetic preferences such as focus area, emotion depth, camera and lighting importance, and stylistic tendencies. Each profile additionally specifies an expertise level and an expression style, which jointly control how the profile manifests as a natural-language prompt during personalized prompt generation. Complete specifications are provided in Appendix~\ref{app:profiles}.
 
%--- Prompt Generation ---
\noindent\textbf{Personalized Prompt Generation.}
\label{sec:prompt_gen}
Given a metadata entry $\mathbf{m}$ and a user profile $\mathbf{u}$, we generate a personalized language video creation instruction $\mathbf{p} = g(\mathbf{m}, \mathbf{u})$ via LLMs.

Four principles govern the generation. First, every prompt must read as a concrete directive naturally typed, with the opening sentence clearly stating the creative intent in a style appropriate to the user's expertise level. Second, higher-priority dimensions receive detailed elaboration while lower-priority ones are mentioned briefly but never omitted, ensuring no evaluation dimension is left ungrounded. Third, different profiles produce stylistically distinct prompts for identical metadata, but the underlying information content remains equivalent. Fourth, nine rejection criteria serve as quality gates, filtering out prompts that lack a clear task statement, omit key dimensions, read as essay-like prose, exceed the character limit, or exhibit expertise-level inconsistencies such as a novice using professional jargon.
 
%------------------------------------------------------------------
\subsection{Dynamic Checkpoint Rubric}
\label{sec:rubric}

A fundamental challenge in evaluating diverse long-form videos with a unified rubric is \emph{checkpoint applicability}: a lip-sync checkpoint is meaningless for a landscape video without characters, while a story-arc checkpoint is irrelevant for a music visualization. To solve this, we introduce a dynamic checkpoint rubric that automatically adapts to each video’s content through a two-stage process. 

% In the first stage, \emph{content profiling}, a vision-language model performs a rapid pass to representative frames and produces a structured Content Profile containing 18 boolean and integer attributes that describe four key aspects of the video: the entities present, the scene structure, the stylistic properties, and the motion characteristics. This profile is further enriched with signals from automatic speech recognition. In the second stage, \emph{checkpoint activation}, we maintain a registry of 40 checkpoints organized into nine sub-metrics across five evaluation dimensions (see Table~\ref{tab:checkpoints}). Each checkpoint is associated with an applicability condition expressed as a logical conjunction over the content attributes. Only checkpoints whose conditions are fully satisfied are activated.
In the first stage, \emph{content profiling}, a vision-language model performs a rapid pass over representative frames and produces a structured Content Profile consisting of 18 semantic attributes (detailed in Appendix~\ref{app:content_profile})
describing entities, scene structure, stylistic properties, and motion characteristics. The profile is further enriched with signals from automatic speech recognition. 
In the second stage, \emph{checkpoint activation}, we maintain a registry
of 40 checkpoints organized into 9 sub-metrics across 5 evaluation
dimensions (detailed in Appendix~\ref{app:checkpoints}). Each checkpoint is associated
with lightweight attribute-based applicability conditions, and only
relevant checkpoints are activated for a given video. This dynamic
routing mechanism avoids evaluating irrelevant criteria and enables more
targeted diagnosis across diverse video scenarios.
% These attributes and evaluation criteria were iteratively designed with
% reference to professional long-form video production criteria. We refined the taxonomy through multiple rounds of pilot
% annotation and workflow analysis to ensure coverage of common long-form
% video failure modes while maintaining practical evaluation cost.

% Activated checkpoints are scored in one of two ways. Likert-style checkpoints use a 1--5 scale, with each level anchored by a detailed textual description following the rubric pattern. These anchors reduce subjectivity and encourage evaluators to use the full range of the scale. Binary checkpoints (pass/fail) are reserved for strictly factual and verifiable criteria, such as object permanence or motion continuity, and support a factual override mechanism in which tool-based evidence can supersede the vision-language model’s judgment when definitive. Each checkpoint also carries a predefined weight for intra-metric aggregation that reflects its relative perceptual importance.
Activated checkpoints are evaluated using rubric-guided ordinal scoring
on a 1--5 scale, where each checkpoint is paired with descriptive
criteria that anchor different quality levels and encourage consistent
assessment across evaluators and video types. Scores are subsequently
normalized to the range $[0,1]$ before aggregation to enable consistent
comparison across metrics and dimensions. 
Considering the differing
perceptual importance of evaluation criteria in long-form video
generation, lightweight aggregation weights are applied within each
sub-metric. These content attributes, evaluation taxonomy, and weighting scheme were 
iteratively designed with
reference to professional long-form video production criteria. We further refined them through multiple rounds of pilot
annotation and workflow analysis to ensure coverage of common long-form
video failure modes while maintaining practical evaluation cost.

%------------------------------------------------------------------
\subsection{Multi-Agent Evaluation Pipeline}
\label{sec:pipeline}

The evaluation pipeline is implemented as a directed acyclic graph (DAG) using LangGraph and is decomposed into four sequential phases. This design supports parallel execution within each phase while enforcing strict dependency ordering across phases.

\noindent\textbf{Phase 0: Preprocessing.}
An orchestrator first extracts rich structured signals from the input video through a tool suite with built-in graceful degradation. The tools compute video probing, shot boundary detection, representative frame extraction, transition analysis, color histogram differences, optical flow, audio separation, and automatic speech recognition. All tool outputs and execution states are packaged into a shared preprocessing output and propagated via a common tool context. Full trace exposure is a distinctive feature of the pipeline: every tool invocation and its result is recorded, allowing both agents and human auditors to inspect the evidential basis of every evaluation decision.

\noindent\textbf{Phase 1: Specialist Evaluation (Parallel).}
Four specialist agents then execute in parallel, each focusing on distinct sub-metrics. The ScriptEvalAgent evaluates
% script reasonableness and novelty. 
narrative coherence, richness, and structural reasonableness.
The VideoEvalAgent assesses visual consistency and quality. The AudioEvalAgent examines background music consistency and narration quality. The StabilityEvalAgent checks generation stability and artifact frequency. All agents inherit from a common base class that automatically activates relevant checkpoints based on the Content Profile, injects tool availability context into prompts, 
% and calibrates confidence scores when upstream tools degrade.
and calibrates confidence using the availability, consistency, and agreement of
intermediate tool outputs and reasoning traces rather than relying on
self-reported model certainty.

\noindent\textbf{Phase 2: Cross-Modal Evaluation.}
After the specialist phase completes, the CrossModalEvalAgent evaluates inter-modality alignment across two sub-metrics: text--video consistency and video--audio synchronization. 
% This agent can directly reference findings from Phase 1 to avoid redundant analysis.
This agent can reference Phase 1 findings as evidence candidates, but verifies cross-modal claims against the metadata and tool outputs.

\noindent\textbf{Phase 3: Diagnosis Synthesis.}
The Diagnosis Synthesizer finally aggregates all checkpoint results into a structured Diagnosis Report. Per-dimension scores are computed as confidence-weighted averages
$S_d = \frac{\sum_{r \in \mathcal{R}_d} s_r \cdot c_r}{\sum_{r \in \mathcal{R}_d} c_r}$
% where $\mathcal{R}_d$ is the set of results for dimension $d$. The profile-weighted overall score is then obtained as
where $\mathcal{R}_d$ denotes the set of activated checkpoint results
associated with dimension $d$. The profile-aware overall score is then
computed as
% $S_{\text{overall}} = \frac{\sum_{d \in \mathcal{D}_{\text{active}}} w_d \cdot S_d}{\sum_{d \in \mathcal{D}_{\text{active}}} w_d}$,
$S_{\text{overall}} =
\frac{\sum_{d \in \mathcal{D}} w_d \cdot S_d}
{\sum_{d \in \mathcal{D}} w_d}$,
where $w_d$ is the user profile's priority weight for dimension $d$. 
% and inactive dimensions are omitted rather than penalized. 
% The stability dimension shares its weight with visual camera quality. 
Beyond numerical scores, the report includes a prioritized bottleneck list, low-confidence flags for human review, radar chart data across five dimensions, and an LLM-generated narrative summary with actionable recommendations tied to low-scoring checkpoints. This checkpoint-level bottleneck localization is a core differentiator of \TheName{}, enabling developers to identify specific failure modes rather than receiving only an opaque aggregate score.

% \textbf{In summary}, by grounding evaluations in objective metadata while incorporating user-specific priorities and constraints, and by exposing the full agent trace including tool invocations and results for accurate confidence scoring, \TheName{} produces rich Diagnosis Reports that provide both quantitative scores and actionable insights into specific bottlenecks across narrative, visual, audio, and cross-modal dimensions. This design enables fine-grained, preference-aware assessment of long-form video generation methods.
% \textbf{In summary}, \TheName{} enables structured, workflow-aware, and
% preference-aware evaluation of generated long-form video through
% checkpoint-level diagnosis across multiple quality dimensions.
%%%%%%%%%%%%%%%%%%%%%%%%%%%%%%%%%%%%%%%%%%%%%%%%%%%%%%%%%%%%
% =============================================================================
% Experiments :: Setup + RQ1 (workflow comparison + bottleneck identification)
% =============================================================================
\section{Experiments and Observations}
\label{sec:experiments}
In this section we use \TheName{} to study three orthogonal questions about long-form video generation. \emph{RQ1} asks how the design of the agentic workflow that drives a generator affects the quality of its output, and what bottlenecks survive current workflows. \emph{RQ2} holds the workflow fixed and asks how much of the remaining headroom the base LLM that orchestrates the agent can recover. \emph{RQ3} holds the generated videos themselves fixed and asks whether personalized evaluation reveals quality structure that a generic single-profile metric collapses.
% To address these outlined research questions, we detail our experimental settings below.

\noindent\textbf{Evaluation Settings.}
All evaluations are run with a fixed multi-agent assessor implemented in \TheName{} (LangGraph DAG).
All evaluator calls share the same LLM backend (default: GPT-5.4~\citep{gpt4}).
The preprocessing toolchain combines ffprobe/ffmpeg video probing and audio extraction, PySceneDetect shot segmentation~\citep{pyscenedetect}, representative-frame extraction, and OpenCV~\citep{bradski2000opencv}-based boundary metrics (SSIM~\citep{ssim}, color-histogram shift~\citep{DBLP:conf/trecvid/MasF03}, optical-flow magnitude~\citep{DBLP:conf/cvpr/WeinzaepfelRHS15}).
For cross-modal grounding, we additionally use MobileViCLIP-Small~\citep{yang2025mobileviclip} for text--video similarity, a lightweight lip-sync proxy~\citep{lip} based on mouth-region motion--audio energy correlation, and Sentence-BERT~\citep{sentencebert} semantic similarity for text--audio alignment; BGM consistency is supported by Librosa~\citep{librosa} feature extraction.
All tool outcomes (success/fallback/failure with latency) are logged and injected into agent prompts to calibrate confidence, and full execution artifacts are saved as per-run debug logs and tool traces for auditability.

\noindent\textbf{Systems under Evaluation.}
We compare four agentic video-generation workflows that differ along a single design axis at a time, so that the resulting per-checkpoint signature can be attributed to a specific design choice. \emph{MovieAgent}~\citep{wu2025automated} is an open-source character-aware planner whose base LLM produces a cast list and per-character beat sheet before delegating to keyframe generators and a text-to-video backbone. \emph{ViMax}~\citep{hkuds2025vimax} is an open-source decompose-stitch pipeline whose base LLM produces a script, then a storyboard, then a first-frame image per shot, then runs per-shot text-to-video, and finally composes background music and voiceover post-hoc. For the closed-source production baselines, we use the \emph{Dreamina Creation Agent}\footnote{\url{https://jimeng.jianying.com/ai-tool/generate}}, an end-to-end agent in which the base LLM compiles each clause of the user prompt into a discrete edit on a private MCP tool chain, and the \emph{Kling Canvas Agent}\footnote{\url{https://kling.ai/canvas/home}}, which lays the storyboard out on a shared visual--audio canvas before generation, so that every shot is rendered with global state in scope.

\begin{figure}[t]
\vspace{-10pt}
  \centering
  \includegraphics[width=\linewidth]{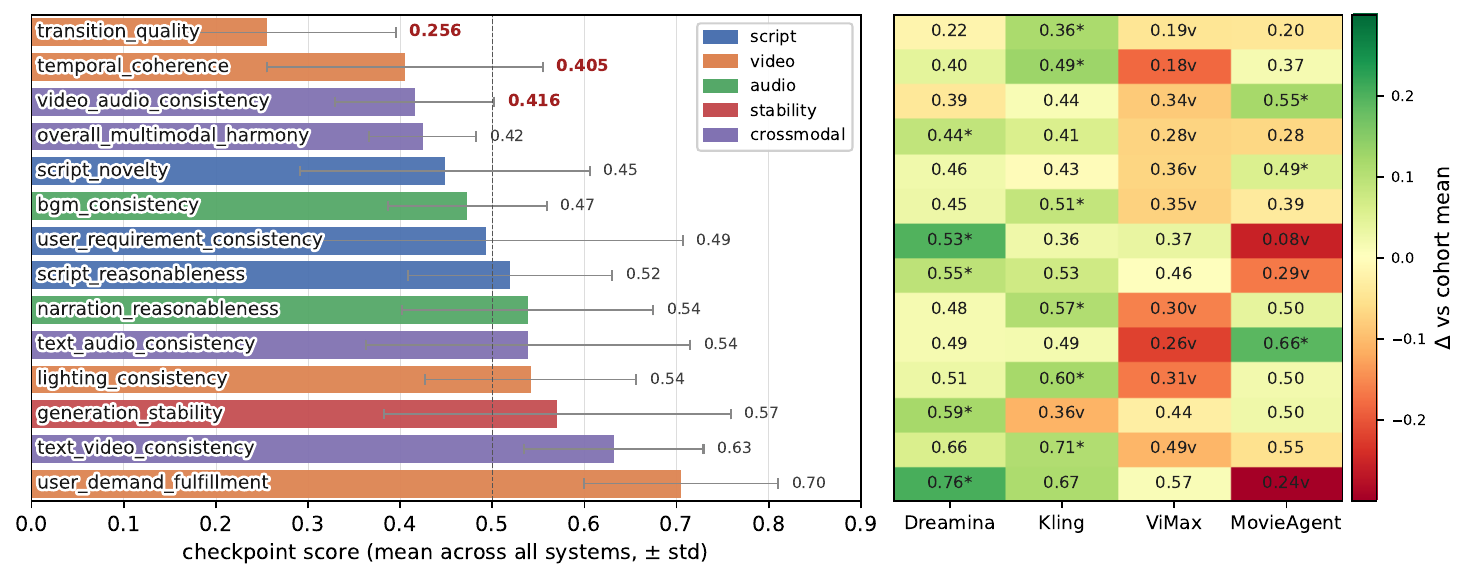}
  \vspace{-15pt}
  \caption{RQ1: Workflow signatures and cohort-wide bottlenecks in long-form video generation. 
  \textbf{Left}: Mean checkpoint scores across all four workflows, sorted ascending, with population standard deviation error bars.
  \textbf{Right}: Heatmap of each workflow’s deviation from the four-workflow cohort mean ($\Delta$). Asterisks mark the cohort-best score per row.
  % Each workflow exhibits a distinct strength--weakness signature traceable to its architectural choices.
  }
  \vspace{-15pt}
  \label{fig_rq1_combined}
\end{figure}

\subsection*{RQ1: How do agentic workflow designs affect final quality, and what are the primary bottlenecks?}

% \noindent\textbf{Approach.}
We operationalize RQ1 in two complementary ways. We first compute the mean of every checkpoint metric on a per-workflow basis and center each row on the four-workflow cohort average, yielding a strength--weakness signature per workflow (right side of Figure~\ref{fig_rq1_combined}). 
% We then pool every latest run from every system and rank the 14 checkpoints by their \emph{global} mean (Left Part of Figure~\ref{fig_rq1_combined}); checkpoints that remain low after pooling are bottlenecks shared by every workflow we test.
We then aggregate results across all
evaluated systems and analyze the globally lowest-scoring checkpoints
(left side of Figure~\ref{fig_rq1_combined}); checkpoints that remain
consistently low are treated as shared bottlenecks
across workflows.

\noindent\textbf{Workflow Signatures.}
Figure~\ref{fig_rq1_combined} reports the workflow-level
strength--weakness signatures across evaluation sub-metrics. The two
closed-source production agents (Dreamina and Kling) outperform the
open-source systems overall, but their strengths arise from contrasting
architectural choices.
Dreamina is strongest on prompt compliance and structural control. Its
edit-action workflow preserves user requirements throughout the
generation process, leading to strong instruction following and stable
generation behavior. However, because edits are applied incrementally
without a persistent global view of the full timeline, the workflow
struggles with transition quality (0.22), revealing a weakness in
cross-shot continuity.
Kling exhibits the opposite profile. Its shared-canvas workflow
maintains stronger global coherence across shots and modalities,
resulting in high text--video consistency (0.71) together with better
temporal and lighting consistency. However, this same global
optimization weakens strict constraint preservation, as detailed prompt
requirements are more easily absorbed into the overall visual
composition.
The two open-source workflows fail for different reasons. ViMax performs
poorly on nearly all coherence-related sub-metrics, especially temporal
coherence (0.18), suggesting that simple decompose-and-stitch generation
is insufficient for maintaining coherent long-form structure. By
contrast, MovieAgent achieves relatively strong audio--dialogue
alignment through character-driven planning, but this abstraction also
suppresses detailed prompt constraints, leading to extremely weak user
requirement consistency (0.08).
% 
% Overall, the workflow signatures reveal that long-form video quality is
% primarily determined by how global structure, constraint preservation,
% and cross-modal state are coordinated throughout the generation process.
Overall, the workflow signatures show that workflow architecture is a
dominant factor in long-form video generation quality, with different
design choices producing distinct trade-offs between global coherence,
constraint preservation, and cross-modal coordination.

\noindent\textbf{Cohort-wide Bottlenecks.}
After pooling all generated videos across workflows, we rank the report-level checkpoint metrics by global mean (Figure~\ref{fig_rq1_combined}). All reported checkpoint and dimension scores are normalized to $[0,1]$. The three lowest-scoring checkpoints share an identifiable structural property: they all probe \emph{between-unit} quality rather than single-frame quality. Transition quality averages $0.256$ across the corpus, temporal coherence averages $0.405$, and video--audio consistency averages $0.416$. Even the cohort-leader on transition quality (Kling, $0.356$) sits well below the $0.5$ midline, so this is not the artifact of a single weak workflow but a property of the underlying generation stack. By contrast the single-frame and prompt-fidelity checkpoints (user demand fulfillment $0.71$, text--video consistency $0.63$, generation stability $0.57$) all sit above the midline.
% \emph{The bottleneck is between units, not within them}: between adjacent shots, and between the visual and audio streams.
The bottleneck is therefore between units rather than within them:
maintaining continuity across adjacent shots and modalities remains the
most persistent weakness across all evaluated workflows.
These results suggest that future workflows may require explicit
global-state propagation or post-generation self-revision mechanisms to
improve cross-shot continuity.

\noindent\textbf{Video-Type Stratification.}
The same bottleneck pattern remains visible after stratifying by
different video types defined in metadata. Across all systems, \emph{Action} is the hardest
category (overall $0.455$), while \emph{Cinematic/Shot-Design}
is the strongest ($0.509$); \emph{Commercial Marketing} and
\emph{Education/News} fall in between ($0.484$ and $0.489$, respectively). The trend is consistent for all workflows. 
% These type-conditioned results reinforce
% the checkpoint-level finding: current agentic workflows are relatively
% robust on composition-driven cinematic tasks, but remain brittle on
% high-dynamics action generation where cross-shot continuity and
% cross-stream synchronization are harder to maintain.
These type-conditioned results further suggest that workflow robustness
depends strongly on how well global state can be maintained under
high-dynamics generation settings, especially for action-heavy videos
requiring sustained temporal and cross-modal consistency.

\subsection*{RQ2: How do base LLM choices affect final quality when the workflow is held fixed?}
\label{sec:rq2}

% \noindent\textbf{Approach.}
RQ2 isolates the effect of the base LLM that orchestrates the agent. We use the model-variation bucket of \TheName{}: the Dreamina edit-action
workflow is held fixed, the same input prompts are shared across all
variants, and only the base LLM that drives the skills and MCP tool-chain
is swapped. Each model is evaluated under the equal-weight neutral
profile.
We evaluate six representative proprietary chat-oriented LLMs~\citep{seed2,glm,claude,gpt4,kimi,minimax}. These models are treated as interchangeable controllers
of the same agentic workflow, allowing us to isolate the effect of the
base LLM from workflow-level components. Detailed descriptions are
provided in Appendix~\ref{app:base_llms}.

\begin{wrapfigure}{r}{0.65\linewidth}
  % \vspace{-10pt}
  \centering
  \includegraphics[width=\linewidth]{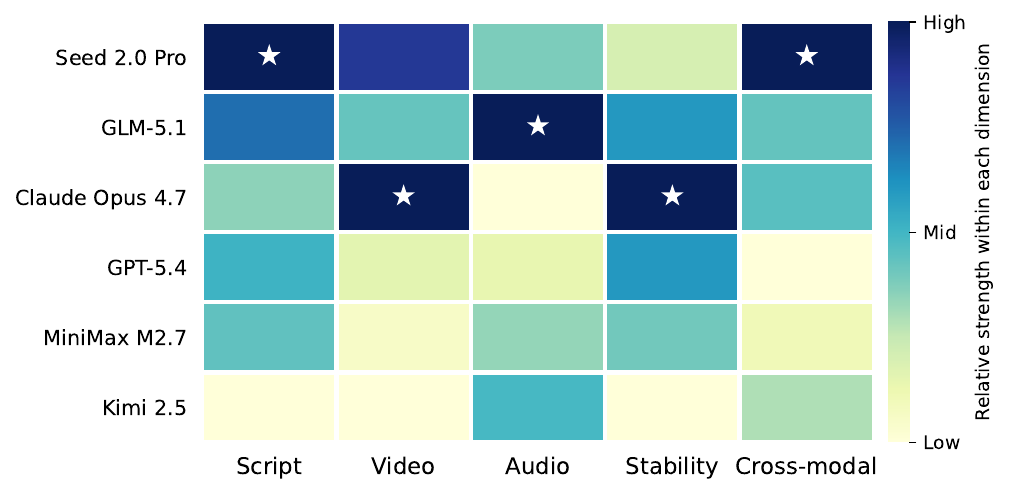}
  \vspace{-10pt}
  \caption{Relative performance across dimensions (column-normalized).
  Darker indicates stronger performance; stars mark per-dimension best.}
  \label{fig:rq2_basemodel}
  \vspace{-10pt}
\end{wrapfigure}

\noindent\textbf{Narrative dimensions dominate the variation across base LLMs.}
Base-LLM choice primarily
affects reasoning-intensive stages, while
leaving system-level components unchanged.
Figure~\ref{fig:rq2_basemodel} shows a column-normalized heatmap of
dimension-level performance. The visual, audio, and stability
% \begin{wrapfigure}{r}{0.65\linewidth}
%   \vspace{-10pt}
%   \centering
%   \includegraphics[width=\linewidth]{fig_rq2_basemodel_relative_heatmap.pdf}
%   \vspace{-10pt}
%   \caption{Relative performance across dimensions (column-normalized).
%   Darker indicates stronger performance; stars mark per-dimension best.}
%   \label{fig:rq2_basemodel}
%   \vspace{-10pt}
% \end{wrapfigure}
dimensions
exhibit low contrast: most models appear in similar color bands,
indicating minimal
differentiation. This confirms that these dimensions
are governed by shared workflow components and are largely insensitive to the choice of base
LLM.
In contrast, the script and cross-modal dimensions display substantially
higher contrast, with clear separation between models. This indicates
that the primary variation across base LLMs lies in narrative reasoning
and multi-modal coordination, which are directly handled by the LLM.

\noindent\textbf{Base LLMs exhibit complementary specialization rather than a strict ranking.}
No single
model consistently dominates across all dimensions. Instead,
models show complementary strengths across
different dimensions
(Figure~\ref{fig:rq2_basemodel}). For example, \emph{Seed 2.0 Pro}~\citep{seed2} leads
on script and
% \begin{wrapfigure}{r}{0.65\linewidth}
%   % \vspace{-10pt}
%   \centering
%   \includegraphics[width=\linewidth]{fig_rq2_basemodel_relative_heatmap.pdf}
%   \vspace{-10pt}
%   \caption{Relative performance across dimensions (column-normalized).
%   Darker indicates stronger performance; stars mark per-dimension best.}
%   \label{fig:rq2_basemodel}
%   \vspace{-10pt}
% \end{wrapfigure}
cross-modal
reasoning, \emph{GLM-5.1}~\citep{glm} on audio,
\emph{Claude Opus 4.7}~\citep{claude} on video and stability, and \emph{GPT-5.4}~\citep{gpt4} on
stability
% \begin{wrapfigure}{r}{0.65\linewidth}
%   \vspace{-10pt}
%   \centering
%   \includegraphics[width=\linewidth]{fig_rq2_basemodel_relative_heatmap.pdf}
%   \vspace{-10pt}
%   \caption{Relative performance across dimensions (column-normalized).
%   Darker indicates stronger performance; stars mark per-dimension best.}
%   \label{fig:rq2_basemodel}
%   \vspace{-10pt}
% \end{wrapfigure}
robustness. The pattern demonstrates that base LLMs specialize in
different aspects of the pipeline rather than forming a strict global
ranking.
From a practical perspective, this implies that model selection can be
task-dependent: applications prioritizing narrative quality favor
script-strong models, while synchronization-sensitive tasks benefit
from cross-modal specialists. 
% However, the magnitude of these differences
% remains modest, and rarely shifts a system from acceptable to
% unacceptable.

% \noindent\textbf{Take-away.}
The above analysis reveals that
% Within a fixed agentic workflow, 
the base LLM acts primarily as a
\emph{narrative and coordination modulator}
within a fixed agentic workflow
. Its impact is concentrated on
script generation and cross-modal alignment, while leaving system-level
dimensions (visual quality, audio fidelity, and stability) largely
unchanged. Compared to the substantial variation induced by workflow
design in RQ1, the effect of base-LLM substitution is secondary. This
suggests a clear optimization priority: improvements should focus on
workflow architecture before model selection, unless the application is
explicitly narrative-sensitive.

% \begin{figure}[t]
%   \centering
%   \includegraphics[width=0.52\linewidth]{fig_rq2_basemodel_relative_heatmap.pdf}
%   \vspace{-4pt}
%   \caption{\small Column-normalized performance across base LLMs
%   (darker = stronger; \* = dimension-best).}
%   \label{fig:rq2_basemodel}
%   \vspace{-8pt}
% \end{figure}

% =============================================================================
% Experiments :: RQ3  (user-profile sensitivity on jimeng_seed2pro)
% =============================================================================
\subsection*{RQ3: Is personalized benchmarking necessary for evaluating user-dependent video quality?}
\label{sec:rq3}

% \noindent\textbf{Approach.}
RQ3 asks whether such a generic protocol is sufficient
when users express different creation preferences.
Under such setting, we hold the metadata
entry, Dreamina agentic workflow, and base LLM fixed, \textbf{but only} vary the user
profile. 
% Figure~\ref{fig:rq3} reflects the combined
% effect of profile-conditioned prompting and profile-weighted evaluation
% under the same underlying metadata intent.

\begin{figure}[t]
\vspace{-10pt}
  \centering
  \includegraphics[width=\linewidth]{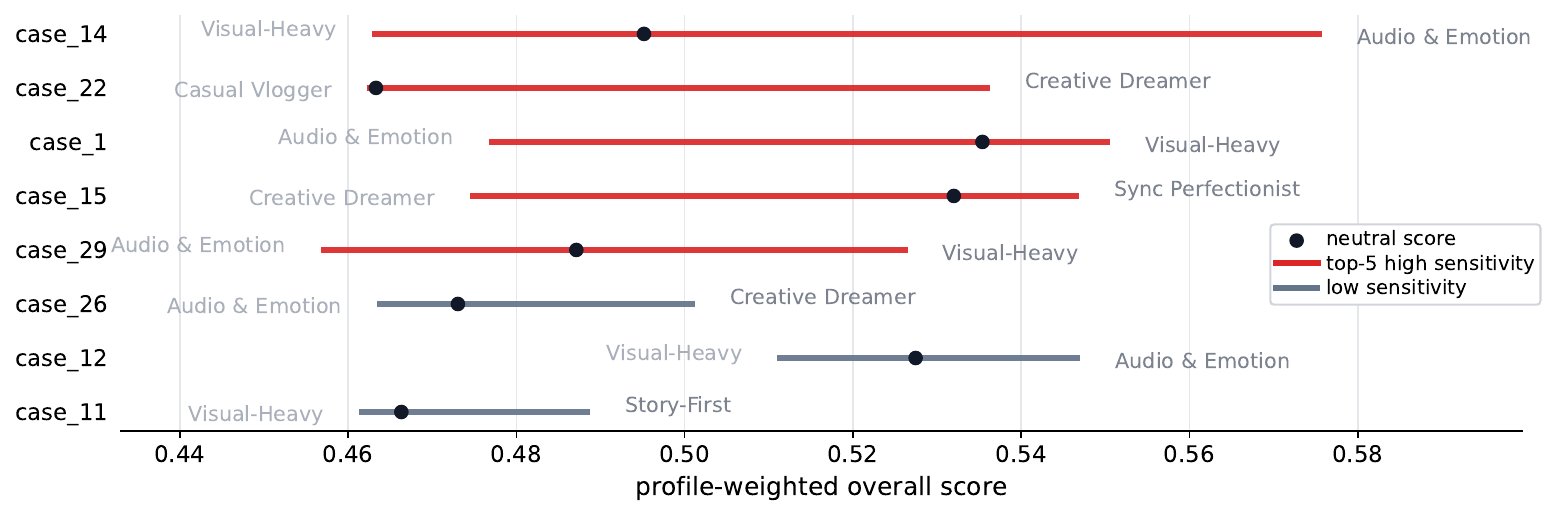}
  \vspace{-15pt}
  \caption{Personalized benchmarking changes the score assigned to the
  same metadata intent. Each segment shows the score range obtained under
  seven user profiles; dots indicate the neutral equal-weight score. We
  show representative high- and low-sensitivity cases, demonstrating that generic evaluation hides substantial user-dependent variation.}
  \vspace{-15pt}
  \label{fig:rq3}
\end{figure}

\noindent\textbf{The same metadata intent leads to different quality outcomes under different profiles.}
Figure~\ref{fig:rq3} shows representative metadata cases, where each
horizontal segment denotes the minimum-to-maximum profile-weighted score
across seven user profiles, and the dot indicates the neutral
equal-weight score. Several cases exhibit wide intervals, with the most
profile-sensitive cases spanning more than ten percentage points. This variation demonstrates that user
profile choice alone can substantially change evaluation outcome with even the same task intent.

\noindent\textbf{Generic evaluation collapses user-dependent quality into a single scalar.}
Under equal-weight aggregation, each case is assigned a single neutral
score, implicitly assuming that all users value dimensions equally.
However, the wide intervals in Figure~\ref{fig:rq3} show that this
assumption hides important profile-specific behavior. A profile that
emphasizes audio and emotional tone may reward a case differently from a
profile that prioritizes visual control or narrative structure. Generic
evaluation therefore averages away the very preference-dependent quality
signals that personalized video creation is meant to capture.

% \noindent\textbf{Take-away.}
In summary,
a workflow that appears strong under equal-weight aggregation may fail
to satisfy specific user intents, while another workflow that is
profile-aligned may be undervalued by the generic metric.
\emph{A single generic score is therefore insufficient to characterize
system performance for content creation scenarios.} Personalized
evaluation provides a necessary complement by exposing how quality varies
with user intent rather than collapsing it into one averaged metric.

%%%%%%%%%%%%%%%%%%%%%%%%%%%%%%%%%%%%%%%%%%%%%%%%%%%%%%%%%%%%
\section{Discussion}
\label{sec:discussion}

The experiments highlight a central challenge in evaluating long-form
video generation: quality is not fully captured by a single scalar score.
Workflow design, base-LLM choice, and user profile affect different parts
of the generation and evaluation pipeline. 
% In particular, our results show
% that persistent failures often arise at boundaries between units rather than within isolated
% frames. 
In this section, we further discuss how these diagnostic signals align with human judgment.

\noindent\textbf{Alignment with Human Judgment.}
To validate whether the diagnostic signals produced by \TheName{} are
consistent with human perception, we conduct an auxiliary human evaluation
with 14 annotators. Each annotator scores the generated videos on a
1--5 Likert scale across 16 fine-grained criteria: character
consistency, scene consistency, object consistency, voice consistency,
style consistency, character-setting plausibility, scene plausibility,
voice plausibility, instruction following,
plot logic, element continuity, segment continuity, camera plausibility,
action quality, duration allocation, and structural stability.
We map these human criteria to the evaluation dimensions used by
\TheName{}.
% Character, scene, object, and style consistency primarily map
% to visual quality and stability; character-setting plausibility, scene
% plausibility, instruction following, and plot logic map to script and
% narrative quality; voice consistency and voice plausibility map to audio;
% multi-character handling, element continuity, segment continuity, camera
% plausibility, action quality, and duration allocation map to cross-shot
% and cross-modal coherence. 
This dimension-level mapping allows us to
compare human judgments with \TheName{} without requiring a one-to-one
match between individual human labels and automated checkpoints.
\emph{Overall, human annotations lead to the same qualitative conclusions as
\TheName{}.} Human raters also identify cross-shot continuity,
segment-level coherence, and multi-modal consistency as the weakest
aspects of current long-form video generation, while prompt fidelity and
local visual quality are judged more favorably. This supports our main
finding that the dominant bottlenecks are between-unit failures rather
than within-shot failures. Human judgments also agree with the motivation
for personalized evaluation: videos that appear acceptable under one
preference emphasis can be judged weaker under another, such as narrative
logic, audio emotion, or visual control. These results suggest that
\TheName{} captures human-perceptible failure modes while providing a more
structured diagnostic decomposition than scalar human ratings alone.

\noindent\textbf{Limitations and Future Work.}
\TheName{} relies on
model-based evaluators and tool outputs, so errors in visual
understanding, speech recognition, shot detection, or intermediate tool
signals can propagate to checkpoint-level judgments. Confidence weighting
and graceful degradation reduce this risk, but future work should further
calibrate the evaluator with larger-scale human annotations.
Meanwhile, \TheName{} focuses on diagnosis rather than automatic
intervention. It identifies where a generation fails, but does not yet
repair the video or select an alternative workflow. A natural next step is
to close the loop: using checkpoint-level bottlenecks to trigger targeted
regeneration, cross-shot repair, audio realignment, or profile-specific
optimization.

%%%%%%%%%%%%%%%%%%%%%%%%%%%%%%%%%%%%%%%%%%%%%%%%%%%%%%%%%%%%
\section{Related Work}
Existing work has developed benchmarks and metrics for evaluating
video-generation systems. For example, VBench~\citep{huang2024vbench} and its successors~\citep{zheng2025vbench2, huang2025vbenchpp}
introduce automatic metrics for subject consistency, motion smoothness,
aesthetic quality, and prompt alignment, while extending evaluation to
more complex dimensions such as controllability and physical realism.
Other efforts, such as Video-Bench~\citep{han2025videobench}, adopt LLM-as-judge pipelines~\citep{li2023llmjudge, zheng2025vbench2} to
assess video quality through question-answering frameworks. However,
these benchmarks are primarily designed for short clips or frame-level
evaluation, and remain limited in capturing long-form video properties
such as multi-shot transitions, narrative coherence, and cross-modal
coordination~\citep{huang2024vbench, zheng2025vbench2, heusel2017gans, salimans2016improved, unterthiner2018towards}. In addition, existing metrics focus heavily on visual
quality and text-video alignment, while audio, storytelling, and
user-dependent preferences are less systematically evaluated~\citep{han2025videobench}. As
long-form video generation increasingly relies on multi-stage and
agentic workflows~\citep{wu2025automated, hkuds2025vimax, meng2025holocine, jang2024mavis, huang2025hollywoodtown}, a benchmark that provides structured, diagnostic, and
personalized evaluation across modalities and pipeline components is
still lacking. In parallel,
recent text-to-video generation methods~\citep{singer2023makeavideo, villegas2023phenaki, kondratyuk2024videopoet} have rapidly expanded video
duration, controllability, and narrative complexity, further increasing
the need for structured long-form evaluation.
Our work addresses this gap by proposing a unified
benchmark for long-form video generation that supports multi-modal,
workflow-aware, and user-aware evaluation.

%%%%%%%%%%%%%%%%%%%%%%%%%%%%%%%%%%%%%%%%%%%%%%%%%%%%%%%%%%%%
\section{Conclusion}

Existing video-generation benchmarks have made important progress in
evaluating visual quality and prompt alignment, but a unified benchmark
for long-form, multi-shot, and agentically generated videos remains
lacking. In this work, we introduce \TheName{}, a personalized
multi-agent diagnostic benchmark that evaluates long-form video
generation across structured metadata, user profiles, and multiple
modalities. Our benchmark provides fine-grained, checkpoint-level
diagnosis across script, visual, audio, cross-modal, and stability
dimensions, enabling more informative evaluation beyond aggregate
scores. Through extensive experiments across workflows, base LLMs, and
user profiles, along with human validation, we demonstrate that
\TheName{} captures meaningful quality differences and aligns with human
judgment.

\bibliographystyle{plainnat}
\bibliography{main}
%%%%%%%%%%%%%%%%%%%%%%%%%%%%%%%%%%%%%%%%%%%%%%%%%%%%%%%%%%%%

\clearpage
\appendix

\section{Metadata Entry}
\label{app:metadata_example}

The metadata entries are structured with JSON documents that encode the complete objective task specification of a target video. Each entry serves as the \emph{ground-truth intent} against which generated videos are evaluated. 

Below is a complete, self-contained example for a narrative video (shortened only for display purposes; the actual benchmark entries contain the full per-shot script, visual specifications, and audio directives). 

\begin{lstlisting}[style=jsonstyle]
{
  "meta_id": "narrative_001",
  "duration_sec": 65.0,
  "video_type": "narrative",
  "main_instruction": "Generate a romantic story of a girl chasing a boy in the rain, featuring a complete three-act arc.",
  "// ======= Core Task Specification (Objective Facts) =======": "",
  "modality_details": {
    "text": {
      "story_arc": {
        "act1_setup": "The girl encounters the boy in the rain",
        "act2_conflict": "A misunderstanding erupts and the boy turns to leave",
        "act3_resolution": "They embrace and reconcile in the rain"
      },
      "script": [
        {
          "shot_id": 1,
          "duration_sec": 15,
          "dialogue": "Wait for me!",
          "narration": "Rain blurs her vision as she runs desperately..."
        },
        {
          "shot_id": 2,
          "duration_sec": 20,
          "dialogue": "",
          "narration": "The misunderstanding hits like a sudden downpour..."
        }
      ],
      "tone_requirements": "romantic_bittersweet, emotional progression"
    },
    "visual": {
      "shots": [
        {
          "shot_id": 1,
          "description": "wide shot of the girl running in the rain",
          "camera_movement": "tracking",
          "lighting": "natural_rainy_glow"
        },
        {
          "shot_id": 2,
          "description": "close-up of tearful eyes",
          "camera_movement": "push_in",
          "lighting": "soft_dramatic"
        }
      ],
      "camera_requirements": ["tracking", "push_in", "orbiting", "handheld"],
      "consistency_requirements": ["character_identity", "clothing", "lighting_shadow", "spatial_layout"]
    },
    "audio": {
      "dialogue": true,
      "lip_sync": true,
      "bgm_style": "soft_piano_orchestral",
      "sound_effects": ["rain_ambient", "footsteps_in_puddle"],
      "tone_control": "emotional_buildup_to_warm_resolution",
      "multi_language": "zh_en_switch"
    }
  }
}
\end{lstlisting}

Below we briefly explain the meaning and purpose of each key.

\begin{itemize}
    \item \textbf{meta\_id}: Unique identifier for this metadata entry (e.g., ``narrative\_001'').
    \item \textbf{duration\_sec}: Target duration of the generated video in seconds.
    \item \textbf{video\_type}: High-level category of the video (e.g., narrative, action, commercial).
    \item \textbf{main\_instruction}: A concise, high-level creative brief that summarizes the overall video concept.
    \item \textbf{modality\_details}: The core block containing detailed requirements for the three modalities (text, visual, audio).
    
    \item \textbf{text}: Narrative and script-related requirements.
      \begin{itemize}
        \item \textbf{story\_arc}: High-level three-act story structure (setup, conflict, resolution).
        \item \textbf{script}: Per-shot script details, including dialogue and narration text.
        \item \textbf{tone\_requirements}: Desired emotional tone and progression of the story.
      \end{itemize}
    
    \item \textbf{visual}: Visual and cinematic requirements.
      \begin{itemize}
        \item \textbf{shots}: Array of per-shot visual descriptions, each with shot content, camera movement, and lighting.
        \item \textbf{camera\_requirements}: List of required camera movement types that must appear.
        \item \textbf{consistency\_requirements}: List of visual consistency constraints (character identity, clothing, lighting, spatial layout, etc.).
      \end{itemize}
    
    \item \textbf{audio}: Audio and synchronization requirements.
      \begin{itemize}
        \item \textbf{dialogue}: Boolean flag indicating whether spoken dialogue is required.
        \item \textbf{lip\_sync}: Boolean flag indicating whether lip-sync is required.
        \item \textbf{bgm\_style}: Desired style and mood of the background music.
        \item \textbf{sound\_effects}: List of required sound effects.
        \item \textbf{tone\_control}: Emotional tone progression for the audio track.
        \item \textbf{multi\_language}: Optional language switching requirements.
      \end{itemize}
\end{itemize}

\section{User Profile Specifications}
\label{app:profiles}

\TheName{} defines seven distinct user profiles that represent different real-world preferences in long-form video creation. Each profile \(\mathbf{u}\) consists of three core components:

\begin{itemize}
    % \item \textbf{Priority weights} \(\mathbf{w} \in \mathbb{R}^4\) (\(\sum_d w_d = 1\)) that control the relative importance of the four main evaluation dimensions: narrative quality, visual cinematography, audio emotion, and cross-modal synchronization.
    \item \textbf{Priority weights} \(\mathbf{w} \in \mathbb{R}^4\) (\(\sum_d w_d = 1\)) that control the relative importance of four user-facing evaluation dimensions: narrative quality, visual cinematography, audio emotion, and cross-modal synchronization. Although \TheName{} evaluates five dimensions in total, the generation stability dimension is treated separately from personalized weighting because it reflects system-level reliability rather than subjective user preference.
    \item \textbf{Hard constraints} that specify non-negotiable requirements. Violation of any hard constraint triggers a significant penalty in the final score for that profile.
    \item \textbf{User taste} descriptors that capture fine-grained aesthetic preferences, expertise level, and expression style, which are used to generate stylistically distinct personalized prompts.
\end{itemize}

The priority weights for each profile are summarized in
Table~\ref{tab:user_profiles}. To improve readability, we abbreviate the
four evaluation dimensions as \textit{Story}, \textit{Visual},
\textit{Audio}, and \textit{Sync}, corresponding to text/story arc,
visual/camera quality, audio/emotion quality, and cross-modal
synchronization respectively.

\begin{table}[ht]
\centering
\small
\caption{Priority weights of the seven user profiles across the four
evaluation dimensions.}
\label{tab:user_profiles}
\setlength{\tabcolsep}{5pt}
\begin{tabular}{lcccc}
\toprule
\textbf{Profile} & \textbf{Story} & \textbf{Visual} & \textbf{Audio} & \textbf{Sync} \\
\midrule
Story-First User   & 0.55 & 0.15 & 0.15 & 0.15 \\
Visual-Heavy User  & 0.15 & 0.50 & 0.10 & 0.25 \\
Audio \& Emotion   & 0.20 & 0.15 & 0.45 & 0.20 \\
Sync Perfectionist & 0.20 & 0.20 & 0.20 & 0.40 \\
Creative Dreamer   & 0.25 & 0.30 & 0.25 & 0.20 \\
Casual Vlogger     & 0.30 & 0.20 & 0.30 & 0.20 \\
Detail Obsessive   & 0.25 & 0.30 & 0.25 & 0.20 \\
\bottomrule
\end{tabular}
\end{table}

Each profile represents a distinct preference pattern commonly observed
in long-form video creation scenarios.

\begin{itemize}
    \item \textbf{Story-First User.} Prioritizes narrative coherence,
    emotional progression, and causal storytelling. This profile values
    strong story arcs and logical structure over complex visual effects
    or cinematic camera movement.

    \item \textbf{Visual-Heavy User.} Emphasizes cinematic presentation,
    including camera movement, lighting quality, and visual aesthetics.
    Minor narrative imperfections are tolerated if the visual experience
    remains compelling.

    \item \textbf{Audio \& Emotion User.} Focuses on emotional impact
    through narration, tone control, and background music. Lip-sync and
    emotional alignment between audio and visuals are treated as
    important requirements.

    \item \textbf{Sync Perfectionist.} Strongly prioritizes cross-modal
    consistency, including lip-sync accuracy, text--visual alignment,
    and synchronized audio--visual transitions.

    \item \textbf{Creative Dreamer.} Prefers imaginative and stylized
    content with unusual camera motion, dreamy transitions, and fantasy
    aesthetics. This profile encourages creative expression over strict
    realism.

    \item \textbf{Casual Vlogger.} Seeks natural, conversational, and
    easy-to-watch content. This profile favors balanced storytelling and
    emotional tone while avoiding overly complex cinematic techniques.

    \item \textbf{Detail Obsessive.} Demands high precision across all
    dimensions, with particular emphasis on lighting consistency,
    synchronization accuracy, and structural coherence throughout the
    video.
\end{itemize}

%%%%%%%%%%%%%%%%%%%%%%%%%%%%%%%%%%%%%%%%%%%%%%%%%%%%%%%%%%%%
\section{Checkpoint Taxonomy}
\label{app:checkpoints}

\TheName{} adopts a hierarchical evaluation taxonomy composed of
dimensions, sub-metrics, and checkpoints. The taxonomy is designed to
support both high-level quality assessment and fine-grained bottleneck
localization for long-form video generation. In total, the benchmark
contains 40 checkpoints organized into 9 sub-metrics across 5 evaluation
dimensions: \emph{Script \& Narrative}, \emph{Visual}, \emph{Audio},
\emph{Cross-Modal}, and \emph{Stability}.

The checkpoint hierarchy was iteratively refined with reference to
common long-form video production and cinematic evaluation practices,
followed by pilot annotation and workflow analysis to ensure broad
coverage of representative failure modes while maintaining practical
evaluation cost. Rather than evaluating all checkpoints uniformly,
\TheName{} dynamically activates only checkpoints that are applicable to
the current video according to the metadata specification and detected
content profile. This design avoids penalizing videos for irrelevant
criteria while enabling targeted diagnosis across diverse generation
scenarios.

Table~\ref{tab:checkpoints_all} summarizes the complete checkpoint registry.
Each sub-metric corresponds to a coherent aspect of video quality, while
individual checkpoints capture finer-grained evaluation criteria within
that aspect.

\begin{table*}[t]
\centering
\caption{Complete checkpoint registry in \TheName{} (40 checkpoints
across 9 sub-metrics and 5 dimensions).}
\label{tab:checkpoints_all}
\small
\setlength{\tabcolsep}{5pt}
\renewcommand{\arraystretch}{1.12}

\begin{tabular}{lll}
\toprule
\textbf{Dimension} & \textbf{Sub-metric} & \textbf{Checkpoints} \\
\midrule

Script \& Narrative
& Script Reasonableness (5)
& \begin{tabular}[t]{@{}l@{}}
event\_chain\_logic \\
character\_motivation \\
pacing\_structure \\
internal\_consistency \\
dialogue\_naturalness
\end{tabular} \\

& Narrative Richness (3)
& \begin{tabular}[t]{@{}l@{}}
premise\_originality \\
narrative\_surprise \\
visual\_creativity
\end{tabular} \\

\midrule

Visual
& Temporal Coherence (7)
& \begin{tabular}[t]{@{}l@{}}
char\_face\_consistency \\
char\_clothing\_consistency \\
object\_permanence \\
background\_consistency \\
scale\_proportion \\
motion\_continuity \\
temporal\_logic
\end{tabular} \\

& Lighting Consistency (4)
& \begin{tabular}[t]{@{}l@{}}
light\_direction \\
shadow\_consistency \\
color\_temperature \\
exposure\_stability
\end{tabular} \\

\midrule

Audio
& BGM Consistency (4)
& \begin{tabular}[t]{@{}l@{}}
bgm\_mood\_match \\
bgm\_transition\_smoothness \\
bgm\_tempo\_pacing \\
bgm\_volume\_balance
\end{tabular} \\

& Narration Quality (3)
& \begin{tabular}[t]{@{}l@{}}
speech\_timing \\
speech\_emotion\_fit \\
speech\_intelligibility
\end{tabular} \\

\midrule

Cross-Modal
& Text--Video Consistency (5)
& \begin{tabular}[t]{@{}l@{}}
scene\_presence \\
scene\_order \\
character\_matching \\
hallucinated\_content \\
setting\_accuracy
\end{tabular} \\

& Video--Audio Synchronization (4)
& \begin{tabular}[t]{@{}l@{}}
lip\_sync\_quality \\
sound\_event\_alignment \\
audio\_continuity \\
ambient\_sound\_match
\end{tabular} \\

\midrule

Stability
& Generation Stability (5)
& \begin{tabular}[t]{@{}l@{}}
visual\_artifact\_frequency \\
resolution\_sharpness \\
temporal\_degradation \\
color\_consistency \\
duration\_completeness
\end{tabular} \\

\midrule
\multicolumn{2}{l}{\textbf{Total}} & \textbf{40 checkpoints} \\
\bottomrule
\end{tabular}

% \vspace{-6pt}
\end{table*}

%%%%%%%%%%%%%%%%%%%%%%%%%%%%%%%%%%%%%%%%%%%%%%%%%%%%%%%%%%%
\section{Content Profiling Attributes}
\label{app:content_profile}

To support dynamic checkpoint activation across diverse long-form video
types, \TheName{} first constructs a lightweight
\texttt{ContentProfile} describing the high-level semantic properties of
the generated video. The content profile is designed to provide
sufficient structural information for checkpoint routing while remaining
computationally efficient for large-scale evaluation.

The current implementation uses a compact schema consisting of 18
attributes and auxiliary signals grouped into five categories:

\begin{itemize}

    \item \textbf{Entity attributes}
    \begin{itemize}
        \item \texttt{has\_characters}
        \item \texttt{character\_count}
        \item \texttt{has\_dialogue}
        \item \texttt{has\_held\_objects}
        \item \texttt{has\_animals}
    \end{itemize}

    \item \textbf{Scene-structure attributes}
    \begin{itemize}
        \item \texttt{scene\_count}
        \item \texttt{has\_scene\_changes}
        \item \texttt{is\_single\_shot}
    \end{itemize}

    \item \textbf{Style/content attributes}
    \begin{itemize}
        \item \texttt{has\_text\_overlay}
        \item \texttt{has\_special\_effects}
        \item \texttt{is\_live\_action\_style}
        \item \texttt{is\_animation\_style}
        \item \texttt{has\_background\_music}
    \end{itemize}

    \item \textbf{Motion attributes}
    \begin{itemize}
        \item \texttt{has\_camera\_movement}
        \item \texttt{has\_fast\_motion}
        \item \texttt{has\_slow\_motion}
    \end{itemize}

    \item \textbf{Auxiliary fusion signals}
    \begin{itemize}
        \item \texttt{vlm\_char\_count}
        \item \texttt{asr\_speaker\_count}
    \end{itemize}

\end{itemize}

Most visual attributes are inferred from a fast VLM pass over
representative shot thumbnails extracted during preprocessing, while
dialogue- and music-related cues are derived from ASR outputs and
separated audio tracks. To improve robustness in multi-character or
multi-speaker scenes, the final
\texttt{character\_count} is conservatively fused as

\begin{equation}
\texttt{character\_count}
=
\max(
\texttt{vlm\_char\_count},
\texttt{asr\_speaker\_count}
).
\end{equation}

The content profile is subsequently used during checkpoint activation to
determine which evaluation criteria are applicable for the current
video. For example, lip-sync checkpoints are activated only when both
speaking characters and dialogue are detected, while narrative
structure-related checkpoints are omitted for purely montage-style or
music-driven videos.

The attribute taxonomy was iteratively designed with reference to common
long-form video production characteristics and refined through pilot
annotation and workflow analysis to ensure broad coverage of practical
evaluation scenarios while maintaining lightweight execution cost.

%%%%%%%%%%%%%%%%%%%%%%%%%%%%%%%%%%%%%%%%%%%%%%%%%%%%%%%%%%%%
\section{Base LLM Details}
\label{app:base_llms}

We include six proprietary large language models (LLMs) as base
controllers in our study. All models are used in their chat-oriented
configuration and interact with the same toolchain under identical
workflow conditions. This setup allows us to isolate the effect of the
base LLM while holding all system-level components fixed.

\smallskip
\noindent
\textbf{Selection rationale.}
The selected models cover a diverse set of commercial LLM families,
spanning differences in training data, alignment strategies, multimodal
capabilities, and inference efficiency. This diversity ensures that the
observed behaviours are not specific to a single vendor or model design,
but reflect broader patterns in how LLMs interact with agentic workflows.

\smallskip
\noindent
\textbf{Model descriptions.}
We include six representative proprietary LLMs as the base controllers
for the agentic workflows evaluated in this study. The selected models
cover a diverse range of reasoning styles, planning capabilities,
tool-use behaviors, and multimodal interaction characteristics. All
models are evaluated under the same workflow setting and evaluation
protocol to isolate the effect of base-LLM substitution.

\begin{itemize}

    \item \textbf{Seed 2.0 Pro~\citep{seed2}.}
    A proprietary multimodal LLM developed by ByteDance, designed for
    reasoning, planning, and content generation in multimedia pipelines.
    It features strong integration with video-oriented tasks and exhibits
    consistent performance in structured generation and cross-modal
    coordination.

    \item \textbf{GLM-5.1~\citep{glm}.}
    A general-purpose LLM from Zhipu AI, optimized for multilingual
    reasoning and structured output generation. It is particularly
    competitive in dialogue and tool-use scenarios, with strengths in
    controllability and format adherence.

    \item \textbf{Claude Opus 4.7~\citep{claude}.}
    A high-capability LLM from Anthropic, emphasizing reasoning depth,
    alignment, and stable long-context behavior. It is designed to
    produce consistent and reliable outputs under complex instructions,
    making it well-suited for multi-step planning tasks.

    \item \textbf{GPT-5.4~\citep{gpt4}.}
    A frontier LLM from OpenAI, optimized for general reasoning, tool
    use, and robust execution across a wide range of tasks. It exhibits
    strong capability in instruction following and error recovery within
    agentic pipelines.

    \item \textbf{MiniMax M2.7~\citep{minimax}.}
    A multimodal-capable LLM from MiniMax, designed for efficient
    inference and interactive generation. It emphasizes
    latency-performance trade-offs and performs well in real-time or
    user-facing generation settings.

    \item \textbf{Kimi 2.5~\citep{kimi}.}
    A chat-oriented LLM from Moonshot AI, with strengths in long-context
    understanding and conversational coherence. It is particularly suited
    for extended dialogue and context-heavy reasoning tasks.

\end{itemize}

\smallskip
\noindent
\textbf{Experimental control.}
All models are accessed via their respective APIs using default decoding
settings. We do not apply model-specific prompt engineering or tuning,
and all models operate under the same prompt templates, tool interfaces,
and workflow execution logic. This ensures that observed differences
reflect intrinsic model behaviour under a controlled workflow, rather
than artefacts of per-model optimisation.

\smallskip
\noindent
\textbf{Scope and limitations.}
We note that all evaluated models are proprietary systems accessed via
remote APIs, and their internal architectures and training data are not
fully disclosed. As such, our analysis focuses on observable behavioural
differences rather than architectural explanations.

\end{document}